# Comparative Study on Supervised versus Semi-supervised Machine Learning for Anomaly Detection of In-vehicle CAN Network

Yongqi Dong[#], Kejia Chen[#], Yinxuan Peng, Zhiyuan Ma

*Abstract*—**As the central nerve of the intelligent vehicle control system, the in-vehicle network bus is crucial to the security of vehicle driving. One of the best standards for the in-vehicle network is the Controller Area Network (CAN bus) protocol. However, the CAN bus is designed to be vulnerable to various attacks due to its lack of security mechanisms. To enhance the security of in-vehicle networks and promote the research in this area, based upon a large scale of CAN network traffic data with the extracted valuable features, this study comprehensively compared fully-supervised machine learning with semi-supervised machine learning methods for CAN message anomaly detection. Both traditional machine learning models (including single classifier and ensemble models) and neural network based deep learning models are evaluated. Furthermore, this study proposed a deep autoencoder based semi-supervised learning method applied for CAN message anomaly detection and verified its superiority over other semi-supervised methods. Extensive experiments show that the fully-supervised methods generally outperform semi-supervised ones as they are using more information as inputs. Typically the developed XGBoost based model obtained state-of-the-art performance with the best accuracy (98.65%), precision (0.9853), and ROC AUC (0.9585) beating other methods reported in the literature.**

## I. Introduction

Among protocols of the in-vehicle network, CAN protocol is the most widely used with its advantages of real-time application of highly integrated data communication, and excellent reliability under error detection. With the development of vehicular networking consisting of massive electronic control units (ECUs) and various interfaces in a more and more complex communication environment, studies had drawn close concerns about vehicular network security. Despite the irreplaceable role and long application history of the CAN protocol in the field of in-vehicle network communication, CAN network has its vulnerabilities: as it is designed with transmission through a fixed frequency without security measures against attacks, CAN is weak in the face of malicious hacking. The vulnerabilities associated with the CAN protocol of in-vehicle networks were highlighted by Wolf et al [1]. With the vast advancement in vehicular automation, the information security of CAN bus is even crucial to the proper operation of automated vehicles or advanced driver assistance systems (ADAS) [2]. To ensure the safety performance of the vehicle, various anomaly detection techniques have emerged in the past decades.

Existing anomaly detection techniques for CAN traffic can be divided into two categories: rule-based methods and data-driven machine learning (ML) methods. Rule-based methods utilize certain features (e.g., frequency) of the CAN message data to identify patterns and develop rules to separate normal and anomaly samples. Hoppe et.al proposed anomaly detection for in-vehicle networks based on message frequency analysis [3]. Taylor et.al developed frequency-based rules utilizing inter-packet timing over a sliding window to identify anomalies [4]. Miller and Valasek proposed a rule-based method analyzing the fraction of message rates to detect anomalous inject fake messages in the CAN bus [5]. In addition to frequency-based methods, Cho and Kang used clock offsets exploiting the intervals of periodic in-vehicle messages to build rules as fingerprints for intrusion detection [6].

In recent years, with the development of ML algorithms and computational power as well as the accumulation of datasets, numerous ML methods had been adopted in the field of anomaly detection. Usually, the data-driven ML methods are applied in a supervised manner, where each instance in the dataset is marked as normal or anomaly. The ML algorithms are trained on the labeled data to fit a model for automatically detecting anomalies on different new input data. ML algorithms can be roughly categorized into traditional ML and deep learning (DL) algorithms. Regarding traditional ML algorithms, decision tree-based anomaly detection is widely used in various fields, e.g., in smart grid [7] and computer network traffic [8]. Tian et.al adopted the Gradient Boosting Decision Tree (GBDT) algorithm for CAN bus intrusion detection and their method delivered a high True Positive (TP) rate and a low False Positive (FP) rate [9]. Recently, with the burgeoning of deep neural network algorithms, many DL based methods were utilized. For example, Hossain et.al proposed an LSTM based system for detecting invalid CAN messages and achieved high accuracy in detecting various attacks [10]. Li et al. proposed a transfer learning approach for intrusion detection of different types of attacks on the Internet of Vehicles (IoV), and their experimental results showed that the model significantly improved detection accuracy when compared to existing ML methods by at least 23% [11]. Mehedi et.al also developed a deep transfer learning mechanism and applied it to four commonly used backbone networks with their proposed deep transfer learning based LeCun Network (LeNet) model obtained the best results [12].

# These authors contributed equally to this work and should be considered as co-first authors.

Yongqi Dong is with Delft University of Technology, Delft, 2628 CN, the Netherlands. (corresponding author; e-mail: y.dong-4@tudelft.nl).

Kejia Chen is with East China Normal University, Shanghai, 200062 China. (e-mail: chenkejia@zju.edu.cn).

Yinxuan Peng is with Cardiff University, Cardiff, CF10 3AT, United Kingdom (email: pengy41@cardiff.ac.uk).

Zhiyuan Ma is with Shanghai Normal University, Shanghai, 201400 China. (e-mail: raphealma@163.com).

Although various data-driven ML methods had been applied to the anomaly detection problem for in-vehicle CAN bus message data, few of them had employed unsupervised or semi-supervised methods. In general, there are always far more normal data than anomaly data. Furthermore, real-world CAN bus anomalous data generation is always related to safety hazards, it can be dangerous to collect enough anomaly data. Also, labeling the dataset would be tedious work. Therefore, it would be more suitable to utilize unsupervised or semi-supervised manners to tackle the problem. Moreover, a comprehensive comparison is required to evaluate the performance of supervised ML versus semi-supervised ML algorithms regarding anomaly detection for in-vehicle CAN bus network data.

To fill the aforementioned research gaps, this study investigated a large scale of CAN traffic data, extracted variable features with effective feature engineering methods, and developed a deep autoencoder based semi-supervised model for anomaly detection. A comprehensive comparison was also drawn regarding supervised ML versus semi-supervised ML algorithms applied for the in-vehicle CAN bus network. A wide range of ML algorithms are incorporated, including both supervised traditional ML (e.g., decision trees, K-Nearest Neighbor algorithm, and XGBoost), deep neural network models (e.g., Long Short-Term Memory, Residual Neural Network, and LeNet), and semi-supervised ML algorithms (e.g., robust covariance, isolation forest, and local outlier factor). Extensive experiments show that supervised ML methods generally outperform semi-supervised ones as they use more information as inputs. Furthermore, the proposed deep autoencoder based model outperformed other algorithms in the semi-supervised group, while the developed XGBoost obtained the best performance beating other state-of-the-art methods reported in the literature.

The main contributions of this paper lie in:

- An effective feature engineering technique is developed and valuable features are extracted.

- A deep autoencoder based semi-supervised machine learning model is proposed to tackle the in-vehicle CAN bus network data anomaly detection problem, which outperforms other semi-supervised learning methods.

- A comprehensive comparison between supervised machine learning and semi-supervised machine learning applied to in-vehicle CAN bus network data anomaly detection is drawn.

- The developed XGBoost based model using the extracted features obtains state-of-the-art performance with the best accuracy precision and ROC AUC beating all other methods reported in available publications.

## II. MACHINE LEARNING FOR IN-VEHICLE CAN-BUS NETWORK DATA ANOMALY DETECTION

### *A. Supervised Machine Learning Models*

Regarding supervised machine learning models applied for in-vehicle can-bus network data anomaly detection, this study examines both traditional ML algorithms (including single classifier and ensemble learning models) and deep neural network based DL models.

#### *1) Traditional ML models*

Traditional supervised ML models can further be categorised into single classifiers and ensemble learning models. Decision trees (DT) and K-Nearest Neighbor (KNN) algorithm are selected single classifiers, while for ensemble learning models, Random Forest (RF) and Extreme Gradient Boosting (XGBoost) are chosen.

Decision trees (DT), as nonparametric supervised methods for classification, learn subgroups within clusters to assault decision boundaries. Training a DT model is a process of finding the optimal rules in each internal tree node in terms of the selected metric. The DT algorithm is the most popular machine learning model which has been widely utilized in the anomaly detection domain [12-14].

The K-Nearest Neighbor algorithm is a basic classification and regression method that classifies an input instance into a class by finding, given a training dataset, the $K$ nearest instances. The KNN method has a high prediction accuracy when processing data by defining a proximity metric between objects. KNN algorithm can deliver good performance in handling different sensor data [13], thus it is selected.

Individual decision tree models are prone to overfitting, therefore ensemble methods are developed. Among various ensemble models, Random Forest and XGBoost are two of the most used ones. RF adopts a bagging technique where the model builds multiple decision trees by bootstrapping a random subset of data points and features for each tree. RF combines several basic estimator predictions to improve its robustness, and each tree in the set is extracted from a sample taken from the training set. RF models have shown notable performance regarding intrusion detection systems [15].

The XGBoost algorithm is derived from the concept of "augmentation", which combines all the predictions of a set of "weak" learners to develop "strong" learners through an additive training strategy. The XGBoost model aims to prevent over-fitting while minimizing computational costs by simplifying the objective functions that allow combining prediction and regularization terms, as well as parallel computations mechanisms during the training phrases. As a variant of the gradient boosted regression tree models, XGBoost is mainly based on the processes of additive learning. To be specific, in the additive learning process, the first weak learner is fitted based upon the input data, then using the residuals of this first learner, a second weak learner is fitted with the objective of reducing the residuals. The ultimate prediction of the model is finally obtained by considering the predictions of each fitted learner. For more details about XGBoost, please refer to [16].

In this study, the Python class *XGBoost* from *xgboost-1.5.2* package, and classes of *DecisionTreeClassifier*, *KNeighborsClassifier*, and *RandomForestClassifier* from *scikit-learn.tree*, *scikit-learn.neighbours*, and *scikit-learn.ensemble* packages respectively were utilized. The to-be-determined parameters in each model, e.g., *max_depth* and *n_estimators* for *XGBoost*, were obtained through training with the *GridSearchCV* function from *scikit-learn.model_selection*.

*2) Deep Learning models*

In recent years, various DL algorithms have advanced and been successfully applied to anomaly detection with excellent performance [12]. This study selected the widely utilized Convolutional Neural Network (CNN), Long Short-Term Memory (LSTM) neural network, together with the deep transfer learning based Residual Neural Network (ResNet) and LeCun Network (LeNet) proposed by Mehedi et.al [12] as the baseline models. To save training time and computational resources, this study cited the results reported in [12] to be compared with other models.

## B. *Semi-supervised Machine Learning Models*

Regarding semi-supervised learning methods, Robust Covariance (RC), Local Outlier Factor (LOF) and Isolation Forest (IF) are selected as the baselines. Furthermore, this study proposed a deep autoencoder (DAE) based semi-supervised model applied for CAN data anomaly detection.

Robust covariance (RC) is a variance-based robust anomaly detection model which assumes that all normal samples obey Gaussian distribution and uses the Mahalanobis distance to derive the outliers.

The Local Outlier Factor (LOF) considers the density of data points in the actual distribution and utilizes local density as a key factor to detect outliers. The locality is calculated by KNN algorithm. LOF algorithm determines the outliers in two dimensions, one is the small reachable density of the target sample points, and the other is the large reachable density of all the K-nearest neighbours of the target samples.

Isolation Forest (IF) is used to separate outliers by continuously cutting subspaces to do unsupervised outlier detection analysis, which is lenient on data features. Usually, the more the number of trees, the more stable the IF algorithm is. And compared to other traditional algorithms such as LOF, IF is robust and has no assumptions about the distribution of the data set. Fei et al. used IF for anomaly detection with excellent performance standing out among many other anomaly detection algorithms [17]. However, it is not suitable for high dimensional data and the accuracy of detection decreases when there are more local outliers.

In this study, the Python classes *EllipticEnvelope*, *LocalOutlierFactor*, and *IsolationForest* from packages of *scikit-learn.covariance*, *scikit-learn*.neighbors, and *scikit-learn.ensemble* were utilized for RC, LOF and IF respectively, and the parameters of each model were obtained through training with the function *GridSearchCV* from *scikit-learn.model_selection*.

*Deep Autoencoder Based Semi-supervised Learning*

This study developed a DAE based semi-supervised learning method for in-vehicle CAN networks message anomaly detection. The framework of the proposed method is illustrated in Figure 1.

The proposed DAE is composed of the encoder and the decoder which are two symmetrical feedforward multilayer neural networks. Under the proposed framework, the input data samples are first fed to the encoder part for feature extraction, through which the compressed feature vector is obtained. Then the decoder decodes the compressed feature vector into the original dimension trying to reproduce the input. To reproduce the input data, the autoencoder must capture the most critical features representing the input data. This is done by using fewer intermediate hidden layer nodes to obtain the compressed feature vector which removes redundant information and keeps the most vital patterns.

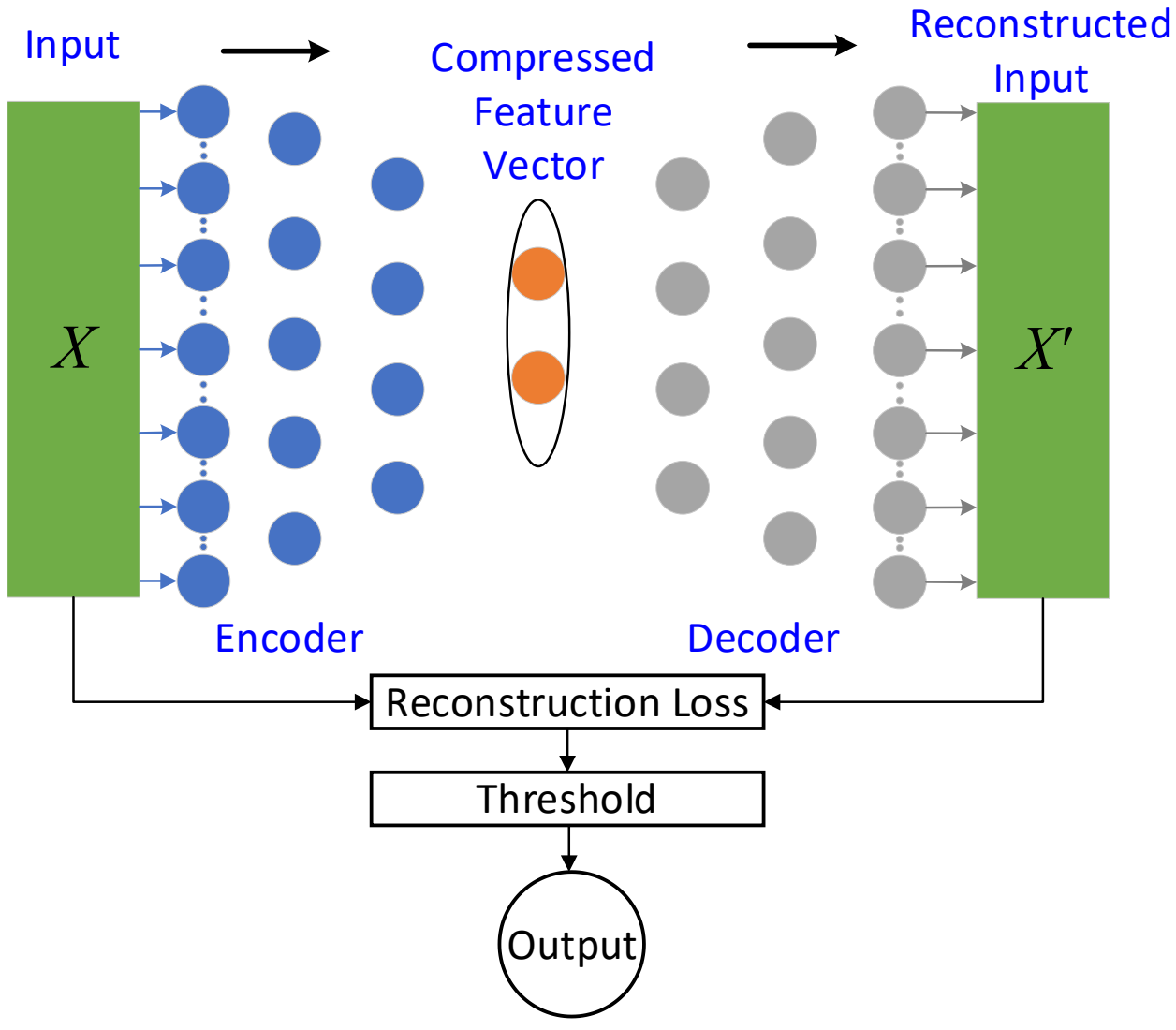

Figure 1. Framework of deep autoencoder based semi-supervised method.

Assume that the input data sample can be denoted by $\boldsymbol{X}$, the compressed feature vector represented by $\boldsymbol{C}$ and the reproduced output represented by $\boldsymbol{X'}$. Then

$$\boldsymbol{C} = g(\boldsymbol{W}\boldsymbol{X} + b) \tag{1}$$

where $g(\cdot)$ is the activation function, $\boldsymbol{W}$ standards for the encoder weights and $b$ is the bias. Thus the expression of $\boldsymbol{X'}$ can be demonstrated by

$$\boldsymbol{X'} = g'(\boldsymbol{W'}\boldsymbol{C} + b') \tag{2}$$

where $g^{'}(\cdot)$, $\boldsymbol{W'}$, and $b'$ are the activation function, weights, and the bias in the decoder part respectively.

The reconstruction losses $\boldsymbol{L}$ can then be calculated by

$$\boldsymbol{L} = l(\boldsymbol{X},\ \boldsymbol{X'}) \tag{3}$$

where $l(\cdot)$ represents the chosen loss function. In this study mean square error (MSE) was selected.

In the training phase, normal data was input to the autoencoder to train the weights with minimizing the reconstruction loss $L$ as the objective and using the backpropagation mechanism as the updating method. Then in the validation phase, a validation dataset that contains a few anomaly data samples is adopted to fine-tune a good threshold. Usually, a good threshold can be expressed by

$$Thrd = \gamma \cdot percentile_p\ \boldsymbol{L} \tag{4}$$

where $percentile_p$ is a function of the $p$th percentile with hyperparameters $p$ and $\gamma \geq 0$.

Finally in the application phase, for a new data sample, after feeding it through the trained autoencoder, its reconstruction loss can be calculated and then compared with the fine-tune threshold $Thrd$. If the loss is beyond the threshold, it will be classified as an anomaly otherwise as a normal one.

## III. EXPERIMENTS AND RESULTS COMPARISON

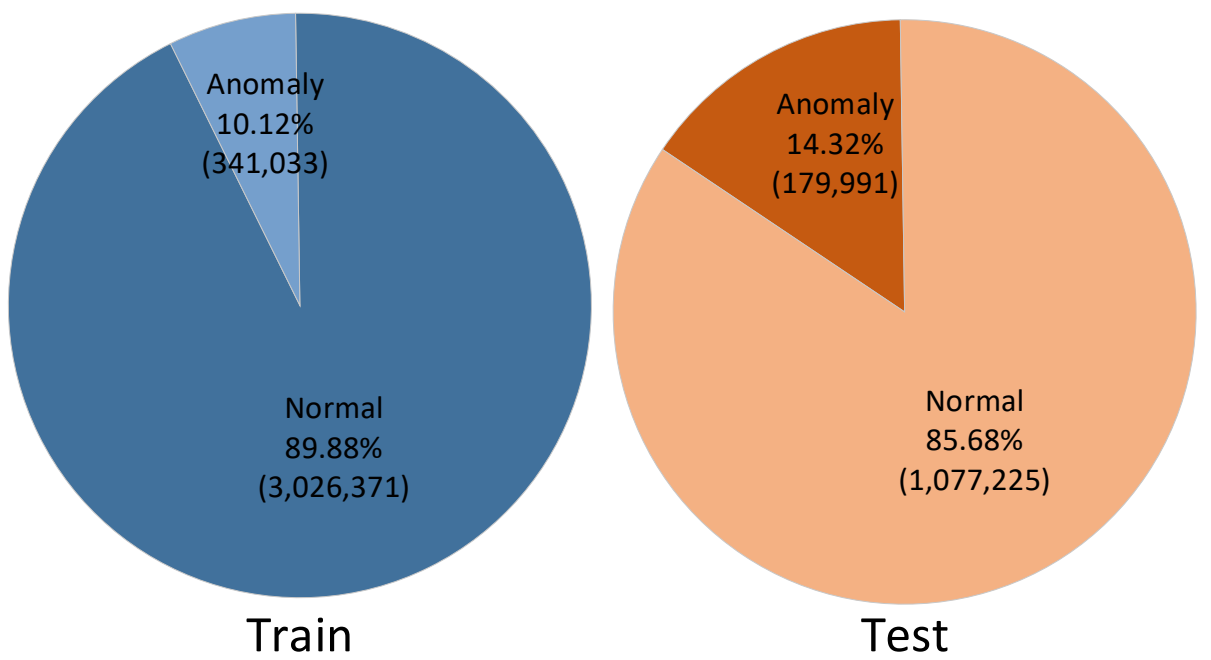

Figure 2. The distribution of the dataset.

### A. *Data Description, Processing and Feature Engineering*

The dataset used in this paper was extracted from CAN network traffic based on the modern Avante CN7 provided by the *Car Hacking: Attack & Defense Challenge 2020* competition [18]. There are two competition rounds, i.e., the *preliminary* round and the *final* round. This study built an integrated large-scale training dataset utilizing the training and submission (which serves for testing) dataset in the *preliminary* round and screened out the data with the status of the vehicle as *stationary* which is the case in the *final* round. The whole dataset in the *final* round was selected as the test set.

Originally, five fields i.e., *Timestamp*, *Arbitration_ID*, *DLC*, *Data*, *Class* are provided based on which this study elaborated 67 features which fall into four categories, i.e., *Arbitration_ID* (CAN identifier, 1 feature), *DLC* (data length code, 1 feature), time interval (1 feature), and data transmitted (CAN data field, 64 features if *DLC* is less than 8 then fills the high-order positions with 0).

For anomaly detection, the used dataset is very sensitive to missing and noisy data with its large size. Originally, there were a total of 3,367,647 instances in the built training dataset and 1,270,310 instances in the testing dataset including noisy and inconsistent data. This study applied various techniques to remove noise and clean inconsistencies of the original datasets, for example, using *dropna* function to remove the instance with *NULL* and missing values. Then, this study converted the *CAN_ID* feature from hexadecimal values into decimal values, and the *Data_Field* feature from hexadecimal values into binary values. Furthermore, a new but very important feature, time interval, was calculated based upon *CAN_ID* and *Timestamp*. To be specific, the data were first grouped by *CAN_ID*. Then, within each subset of the same *CAN_ID*, data samples were increasingly sorted by *Timestamp*. The time interval was obtained by calculating the increasing *Timestamp* value of the adjacent later sample minus the former one.

Finally, a total number of 3,367,404 samples were obtained as the training set with 3,026,371 normal instances and 341,033 anomaly instances; while a total number of 1,257,216 samples were obtained as the testing set with 1,077,225 normal instances and 179,991 anomaly instances. The distribution of normal and anomaly instances within the training set and testing set is illustrated in Figure 2.

### B. *Evaluation Metrics*

Various metrics are used to evaluate the overall performance of the selected model. Four basic terms, i.e., True-positive (TP) which represents the number of correctly detected anomalies, True-negative (TN) which represents the number of correctly detected normals, False-positive (FP) which represents the number of incorrectly detected anomalies, and False-negative (FN) which represents the number of incorrectly detected normals, are first obtained. Then, based on the four terms, accuracy, precision and recall were calculated.

Accuracy is the percentage of correctly predicted samples in the total sample, whose mathematical expression can be defined as follows:

$$Accuracy = \frac{TP+TN}{TP+TN+FP+FN} \tag{5}$$

Precision is the number of correctly predicted positive anomaly observations as a percentage of the number of predicted positive anomaly observations and shows how close the measurements are to each other. The mathematical expression of precision is defined by

$$Precision = \frac{TP}{TP+FP} \tag{6}$$

Recall ratio is the percentage of positive observations correctly predicted in the actual category.

$$Recall = \frac{TP}{TP+FN} \tag{7}$$

Finally, the F1-score provides an overall view of recall and precision (weighted average). F1-score ranges from 0.0 to 1.0, with 1.0 indicating perfect precision and recall.

$$F1\text{-}score = 2 \times \frac{Precision \times Recall}{Precision+Recall} \tag{8}$$

Another significant indicator for evaluating the two-class classification problem is the Receiver Operating Characteristic-Area Under the Curve (ROC AUC) which determines areas where the evaluated model is classified better within normal and anomaly situations. To measure ROC AUC, one needs diagnostic accuracy, which depends sensitivity expressed by the true positive rate (TPR), i.e., recall ratio, and the specificity expressed by the true negative rate (TNR). TPR and TNR are demonstrated by the following two equations

$$TPR = \frac{TP}{TP+FN} \tag{9}$$

$$TNR = \frac{TN}{TN+FP} \tag{10}$$

### C. *Results Comparison*

Table I synthesizes the quantitative performance comparison results of the selected models. As shown in the table, supervised methods generally outperform semi-supervised methods which can be explained by the truth that they utilize more information (i.e., the true labels of the data) as the inputs. Typically, XGBoost algorithm in the supervised category shows the optimal performance among all compared algorithms, with the best Accuracy (98.65%), Precision (0.9853), and ROC AUC (0.9585). These results also beat the best score reported in the *Car Hacking: Attack & Defense Challenge 2020* [18]. To better visualize the results of XGBoost, Figure 3 demonstrates its confusion matrix from which one can identify the correctly classified number of normal and anomaly instances.

TABLE I. PERFORMANCE COMPARISON OF THE MODELS

| Algorithms | Accuracy | Precision | Recall | F1-score | ROC AUC |
|---|---|---|---|---|---|
| ***Supervised Methods*** | | | | | |
| DT | 0.9177 | 0.6591 | 0.8805 | 0.7539 | 0.9022 |
| DT* | 0.9532 | 0.9463 | 0.9558 | 0.9608 | 0.9408 |
| KNN | 0.9254 | 0.7827 | 0.6631 | 0.7180 | 0.8162 |
| KNN* | 0.9248 | 0.9141 | 0.9390 | 0.9215 | --- |
| RF | 0.8497 | 0.4859 | 0.8576 | 0.6204 | 0.8530 |
| RF* | 0.9448 | 0.9448 | 0.9245 | 0.9216 | --- |
| XGBoost | **0.9865** | **0.9853** | 0.9193 | 0.9512 | **0.9585** |
| XGBoost_3** | 0.9697 | 0.9780 | 0.8062 | 0.8838 | 0.9016 |
| XGBoost_66** | 0.9216 | 0.9231 | 0.4934 | 0.6430 | 0.7432 |
| CNN* | 0.9590 | 0.8913 | 0.7891 | 0.8852 | 0.9209 |
| LSTM* | 0.9762 | 0.9808 | 0.9392 | 0.8844 | 0.9288 |
| ResNet* | 0.9795 | 0.8958 | 0.8845 | 0.9001 | 0.9703 |
| LeNet* | 0.9810 | 0.9814 | **0.9804** | **0.9783** | 0.9542 |
| ***Semi-supervised Methods*** | | | | | |
| RC | 0.7523 | 0.1192 | 0.1142 | 0.1166 | 0.4866 |
| LOF | 0.5358 | 0.1595 | 0.5252 | 0.2447 | 0.5314 |
| IF | 0.8285 | 0.3133 | 0.1662 | 0.2172 | 0.5527 |
| DAE | 0.8752 | 0.8503 | 0.1558 | 0.2634 | 0.5756 |

*. Results reported in [12]

**. Results for ablation study.

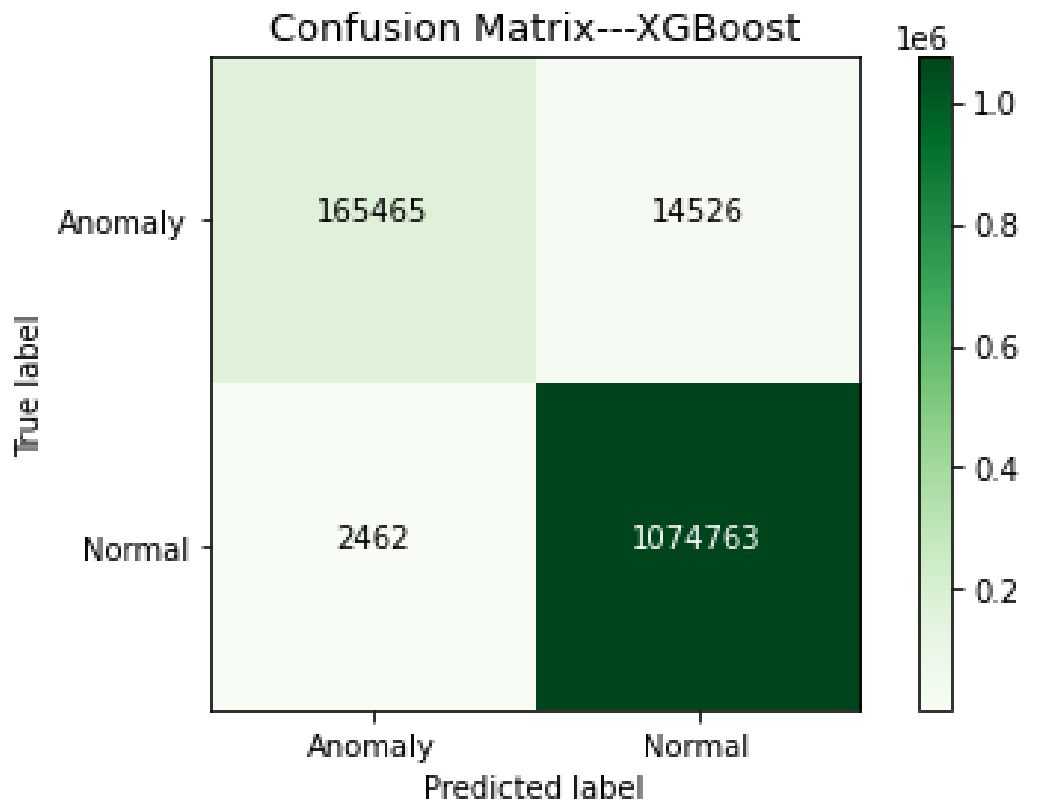

Figure 3. The confusion matrix of XGBoost.

Furthermore, for the same algorithms, the results reported in [12] are always better than the results obtained in this study. The reason is that, in [12], only the data in the final round of the *Car Hacking: Attack & Defense Challenge 2020* [18] are utilized, which means that training and testing datasets are from the same dataset. While in this paper, the testing set is a completely new dataset that is not shown in the training process, plus both the train and test set possess far more instances compared with those in [12] which are only 1,005,843 for training and 251,460 for testing respectively.

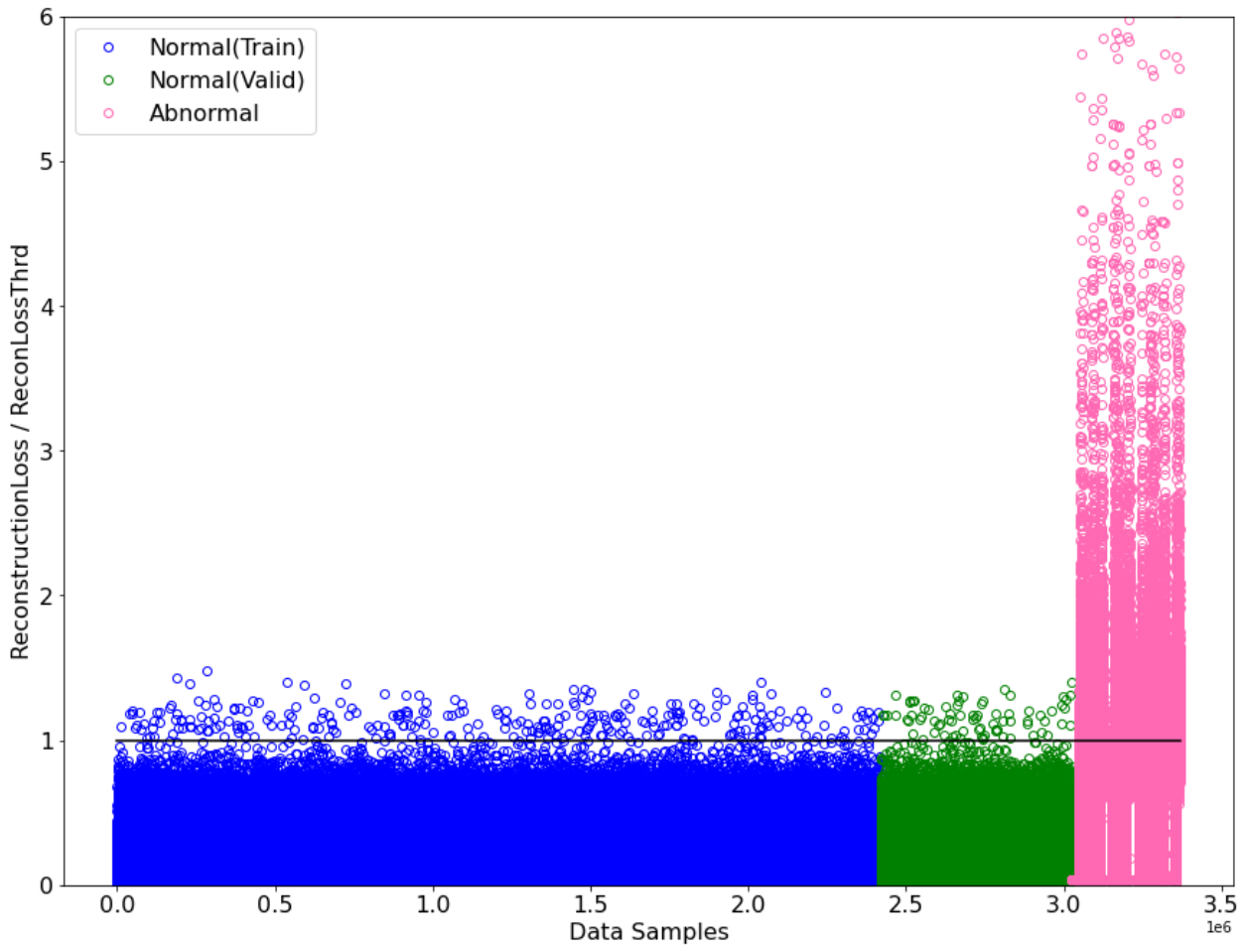

Figure 4. The visualization for reconstruction loss of the deep autoencoder based semi-supervised learning method.

From Table I, it is also noticed that, although all the semi-supervised methods do not behave well, the proposed DAE based semi-supervised learning method obtains the best performance among the ones in the semi-supervised group. Figure 4 visualizes the reconstruction loss obtained from the DAE based semi-supervised learning method. It can be seen that a large proportion of the anomaly data instances will obtain higher reconstruction losses compared with the threshold, however, there are still anomaly data samples that deliver very low losses which fail the algorithm. These findings somehow verify the potential effectiveness of the proposed DAE method. The reason why semi-supervised learning methods do not work well in this CAN network anomaly detection use case might be related to the properties of the data which need further exploration in future studies.

## D. *Ablation Study*

To further verify the effectiveness of the elaborated feature engineering method, the extracted valuable features, and the best method of the developed XGBoost model, detailed ablation studies were carried out. Figure 5 illustrates the relative importance of the utilized features when using XGBoost, in which the first 64 features are data transmitted, and the last three features are *DLC*, *Arbitration_ID*, and time interval respectively. One can identify that the most important feature is the proposed newly calculated time interval, while the data transmitted expressed in binary values are also important. However, these valuable features are not covered in the previous study using similar datasets [12].

Furthermore, in Table I, three XGBoost model variants are used for the ablation study, in which XGBoost_3 means the XGBoost model using only the last 3 features, while XGBoost_66 means the XGBoost model using the first 66 features without time interval. It is witnessed that without time interval, all evaluation metrics of XGBoost_66 decrease compared with XGBoost using the full 67 features, and even XGBoost_3 outperforms XGBoost_66 which further demonstrates the importance of the proposed time interval feature.

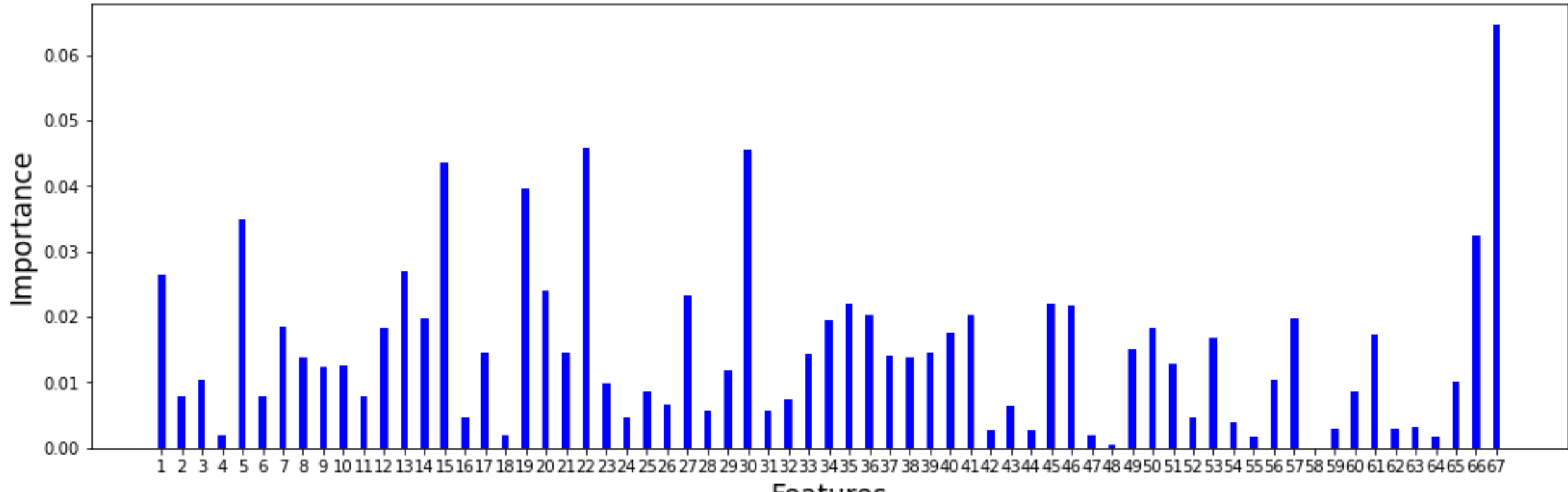

Figure 5. The relative importance of the utilized features under XGBoost method.

## IV. DISCUSSION AND CONCLUSION

In the era of burgeoning technologies of automated vehicle and ADAS systems, the security of in-vehicle CAN network bus is more and more crucial to the safety and security of vehicle driving. To enhance the security of CAN bus networks and promote research on anomaly detection in related areas, this paper examined a large scale of open-sourced CAN network traffic data and extracted valuable effective features, implemented and tested various ML models. As few studies employed semi-supervised learning methods, to fill the research gap, this study proposed a DAE based semi-supervised model using the reconstruction loss as a measure to detect potential anomalies. The proposed DAE based method obtained the best performance among all the algorithms in the semi-supervised category. From the visualization of the reconstruction losses for both normal and anomaly samples, the potential effectiveness of the proposed semi-supervised DAE method can be somehow verified, although further explorations need to be investigated.

Comprehensive comparisons were carried out comparing fully-supervised versus semi-supervised ML methods applied for CAN message anomaly detection. Various traditional ML models, e.g., DT, KNN, XGBoost, and state-of-the-art deep neural network models, e.g., ResNet and LeNet, are evaluated. Extensive experiments on the large-scale dataset demonstrate that the fully-supervised methods generally outperform semi-supervised ones as the labeled information is utilized in inputs which is surely very helpful. Therefore, one valuable insight is that although, in reality, it is difficult to obtain labels for anomalies, if the relevant labels are available, it is better to make full use of them through fully supervised learning methods. Furthermore, the developed XGBoost based model obtained state-of-the-art performance with the best accuracy, precision, and ROC AUC, beating all other methods reported both in this paper and in related literature using the same or similar dataset.

The reason why semi-supervised machine learning methods do not work well in this use case might have something to do with the characteristics of the utilized CAN traffic data. However, further explanations need to be explored in future studies.